# Machine Learning and Optimization Techniques for Solving Inverse Kinematics in a 7-DOF Robotic Arm

Enoch Adediran
School of Science, Engineering and Environment
*University of Salford*
Manchester, United Kindom
e.m.adediran@edu.salford.ac.uk

Salem Ameen
School of Science, Engineering and Environment
*University of Salford*
Manchester, United Kindom
s.a.ameen1@salford.ac.uk

**Abstract**

As the pace of AI technology continues to accelerate, more tools have become available to researchers to solve longstanding problems, Hybrid approaches available today continue to push the computational limits of efficiency and precision. One of such problems is the inverse kinematics of redundant systems. This paper explores the complexities of a 7 degree of freedom manipulator and explores 13 optimization techniques to solve it. Additionally, a novel approach is proposed to contribute to the field of algorithmic research. This was found to be over 200 times faster than the well-known traditional Particle Swarm Optimization technique. This new method may serve as a new field of search that combines the explorative capabilities of Machine Learning with the exploitative capabilities of numerical methods.

## I. Introduction

A. Motivation

There is no such thing as a free lunch. This is according to Wolpert and Macready [1] who argued that without re-sampling, all optimization methodologies exhibit equiprobable performance averages. The No Free Lunch Theorem of Optimisation (NFLT) opposes the idea of a one-size-fits-all optimisation technique and posits that one method is preferable to another only when tailored to the specific search space. Knowing this, this case study aims to compare and optimize existing numerical methods, heuristic algorithms, and machine learning techniques. The challenge involves solving a 7 degree of freedom (7-DOF) manipulator's Inverse Kinematics (IK) using AI and various optimization methods including a novel technique. However, the same methods may work for other hyper-redundant structures and robotics disciplines.

B. Complexities of IK in 7-DOF

Serial manipulators with 6 degrees of freedom are extremely popular since this the minimum required to reach a volume of 3D space from any given orientation. Nonetheless, they tend to run into singularities and are susceptible to gimbal locking. 7-DOF are therefore more attractive since they do not have this problem. In addition, their increase in manipulability provides more flexibility in joint limits and obstacle avoidance [2]. The global, closed form solution of a 6-DOF is solvable, and has been shown to consist of a set of polynomial equations with a tight upper bound of 16 solutions [3]. However, the extra DOF of the kinematic structure creates an infinite number of joint configurations as a solution manifold in the joint space, leading to the highly researched field of redundancy resolution.

The classic approach to resolving redundancy uses damped least squares, taking the pseudo-inverse of the Jacobian with local null-space optimization. This has been shown to be robust and computationally efficient at the joint velocity level [4]. It resolves redundancy at the velocity, acceleration and torque domain, but not the position domain [5].

Most efforts towards finding a global, closed form solution for redundant problems in the position domain usually involve the use of linearized first-order instantaneous relations after first mapping from the velocity domain [6], or the utilization of additional position constraints. Several parameterization techniques have been developed including those of the task space [7], joints [8], elbow twist [9], and arm angles [10]. With regards to an analytical approach, these techniques effectively reduce the DOF and transform the equations into closed forms. However, given an algebraic variety of feasible IK solutions, among the literature, an algebraic computation issue has largely evolved into an optimisation problem instead [11].

C. Literature Review

The study of IK optimisation methods for 7-DOF manipulators is still an active and developing subject of research. A review of the recent literature categorizes existing methods into four main groups: analytic, iterative, data-driven and hybrid approaches.

Analytical methods either seek to derive closed form solutions through algebraic manipulations or exploit the geometric properties inherent in the kinematic structure. Chou and Liu [12] introduced an analytical approach employing a novel arm angle parameterization method. Their proposed solution showed high computational speed and exceptional accuracies while avoiding singularities and the joint limits of their 7-DOF arm.

Iterative methods refine an initial guess through a series of steps until a satisfactory error threshold is met. They include numerical techniques such as root-finding Newton-based, and minimization methods. Woliński and Wojtyra [13] demonstrated a solution for the "KUKA LWR 4+" 7-DOF manipulator employing predictive quadratic programming, showing improvement implications for trajectory scaling. Xu, et al. [14] combined the high numerical stability of the damped least-squares with precision capabilities of the Newton-Raphson method to produce an algorithm that is performs optimally, regardless of the initial end effector conditions.

Iterative methods also include heuristic and metaheuristic methods. Heuristic methods tend to have lower computational costs when performing iterations. The 2 most common types are the Cyclic Coordinate Descent (CCD) [15] and the Forward and Backward Reaching Inverse Kinematics (FABRIK). Both FABRIK [16] and CCD [17] have been successfully demonstrated on 7-DOF structures.

Some metaheuristic algorithms that have been successfully applied to the inverse kinematics optimization of 7-DOF serial manipulators include Genetic Algorithms (GA) [18], Differential Evolution (DE) [19], and Artificial Bee Colony (ABC), among several others. The general trend in the literature seems to indicate that the swarm-based types perform particularly well for this problem domain.

One of the most widely used swarm-based methods are the Particle Swarm Optimization (PSO) and its variants. One of such variants is a Quantum PSO which was applied by Dereli and Köker [20] in their 7-DOF model. Wu, et al. [21] successfully integrated an Improved Artificial Bee Colony algorithm with an unconstrained optimization model based on the quaternion method for their 7 DOF manipulator. Their method also outperformed the classic ABC and PSO versions.

Data driven methods employ statistical and machine learning techniques to learn mappings between the desired end effector positions and joint configurations. Artificial Neural Network (ANN) techniques while promising, have yet to demonstrate the same level of positional accuracy resolution as the other methods mentioned when used on their own. Alebooyeh and Urbanic [22] built an ANN architecture for the 7-DOF "YuMi IRB 14000" robot and achieved an accuracy of about $10^{-1}$ mm. Theofanidis, et al. [23] had a similar approach with the 7-DOF Sawyer Robotic Arm using 5 hidden layers and achieved a slightly better accuracy in the range of $10^{-2}$ mm. However, other techniques like the Support Vector Regression (SVR) have been shown to be more effective than ANN [24]. Overall, these data driven methods can achieve good accuracies while maintaining fast processing speeds.

Hybrid methods combine different techniques to leverage their strengths and mitigate their weaknesses.

D. Paper Outline

This paper is organized as follows. Section II introduces the robot arm kinematics and presents each of the optimization algorithms that was used. Section III explores the process of hyperparameter tuning for each of the algorithms and the results from the tests that were performed. Section IV contains the conclusions.

## II. Methodology

A. The Robot Arm Case Study

1. Kinematic Model

The robot arm used in this case study is based on the research published by Ricardo Xavier Da Silva, et al. [25] which presented an IK solution of the "Kuka LBR iiwa 7 R800" cobot using the Grobner Bases Theory.

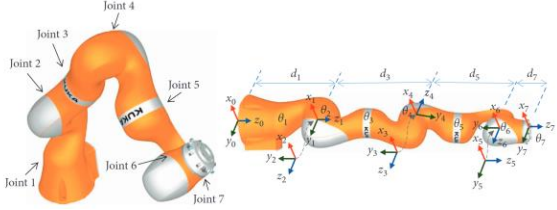

**Figure 1**. Coordinate Frames for the cobot [25]

**Table 1**. DH parameters for the robot

| i | $\alpha_i$ (rad) | $a_i$ (mm) | $d_i$ (mm) | $\theta_i$ (rad) |
|---|---|---|---|---|
| 1 | $-\pi/2$ | 0 | $d_1$ | $\theta_1$ |
| 2 | $-\pi/2$ | 0 | 0 | $\theta_2$ |
| 3 | $-\pi/2$ | 0 | $d_3$ | $\theta_3$ |
| 4 | $\pi/2$ | 0 | 0 | $\theta_4$ |
| 5 | $-\pi/2$ | 0 | $d_5$ | $\theta_5$ |
| 6 | $\pi/2$ | 0 | 0 | $\theta_6$ |
| 7 | 0 | 0 | $d_7$ | $\theta_7$ |

The forward kinematic equations can be gotten by dot multiplying the homogenous transformation matrices for each of the joints.

$$^0T_7 = {^0T_1} \cdot {^1T_2} \cdot {^2T_3} \cdot {^3T_4} \cdot {^4T_5} \cdot {^5T_6} \cdot {^6T_7} \quad (1)$$

The final 4x4 homogenous transformation matrix was derived which provide the algebraic equations for our FK.

$$^0T_7 = \begin{bmatrix} ^0R_7 & ^0P_7 \\ 0 & 1 \end{bmatrix} \quad (2)$$

Where $^0R_7$ is a 3x3 matrix of the Rotation component and $^0P_7$ is a 1x3 matrix that describes the x,y and z position vector from the base to 7th frame. These derived equations were cross-checked with the paper and found to be accurate.

2. Robot Workspace

The serial manipulator has a 7R joint configuration. The equation of a sphere with radius R would satisfy the equation:

$$x^2 + y^2 + z^2 = R^2 \quad (3)$$

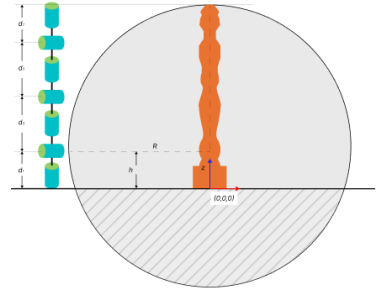

**Figure 2**. Task Space of the 7-DoF redundant manipulator

Since the reference point is from the centre of the base, means that we would apply an offset to z. Therefore, a random point $(x, y, z)$ should fall within the sphere with centre $(0, 0, h)$ which would satisfy the equation,

$$x^2 + y^2 + (z-h)^2 \leq R^2 \quad (4)$$

where $h = d_1$ and $R = d_3 + d_5 + d_7$

Since the point density is inversely proportional to the square of the radius, evenly spaced target positions could be generated using the square root of a random value between zero and $R^2$.

$$(x, y, z) = \sqrt{random} * \cos(angle) \quad (5)$$
$$\text{where } 0 < random < R^2$$

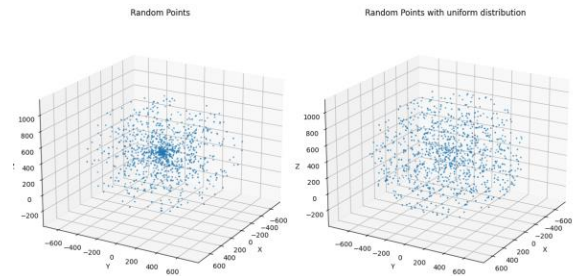

Figure 3. Uniform vs Nonuniform random generation

B. Optimization Techniques.

14 methods in total were evaluated.

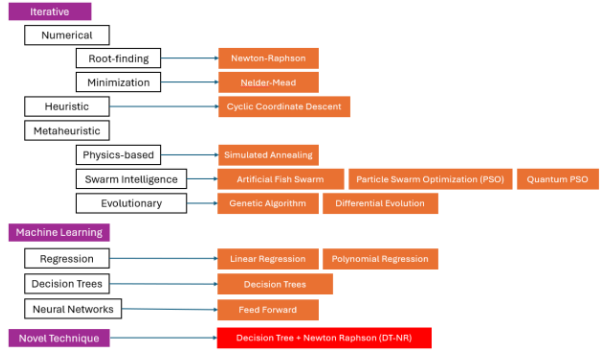

**Figure 4**. Optimization methods used.

Numerical methods in solving the inverse kinematics problem use mathematical techniques to iterate towards approximate solutions. Given a system of equations f(x), root-finding methods are used to find the values of the input variables (roots) such that f(x) = 0. Similarly, minimization methods are used to find the input values that minimize (or maximize) the output of a scalar function.

Heuristic methods in the context of IK include examples such as the Triangulation IK, Cyclic Coordinate Descent (CCD) or Forward And Backward Reaching Inverse Kinematics (FABRIK) [26]. These methods are characterized by their relatively straightforward mathematical formulation, rendering them less prone to convergence issues and obviating the need for Jacobian matrices or derivative computations. Nonetheless, they exhibit higher iteration counts before reaching convergence towards a solution.

Graphs were plotted for the iterative methods, and their averages recorded. A fitness function $f(\vec{x})$ was developed to evaluate positional errors and facilitate the convergence of the output. It does this by returning the linear distance from the end effector to the desired target position as a float. Unlike a multi-objective or dynamic optimization problem, this is a static optimization problem as the goal is to find the optimal set of joint angles.

C. Numerical Methods

1. Newton-Raphson (NR)

The Newton-Raphson method is one of the most widely used root-finding method. It starts with an initial guess, $x_n$ and a smooth, continuous function $f(x)$. It then iterates towards a new guess, $x_{n+1}$ which is where the slope of the tangent line meets the x axis ($f(x) = 0$). The new guess $x_{n+1}$ is given by the following recursive equation.

$$x_{n+1} = x_n - f(x_n)/f'(x_n) \quad (6)$$

In the case of our multidimensional IK problem, our goal is to get a 7x1 Jacobian matrix of the objective function f(a) with respect to the joint angles:

$$\theta = [\theta 1, \theta 2, \theta 3, \theta 4, \theta 5, \theta 6, \theta 7]$$

Each element of the Jacobian is computed as a partial derivative of the objective function with respect to $\theta_i$.

$$J = \frac{\partial F}{\partial \theta} \quad (8)$$

We the update the joint angles iteratively using the formula.

$$\theta_{new} = \theta_{old} - (J^+ \Delta e) \quad (9)$$

Where $J^+$ is the pseudo-inverse of the Jacobian matrix and $\Delta e$ is the difference between the current end position and the target position.

2. Nelder-Mead Algorithm

Two minimization methods were tested, the Constrained Optimization BY Linear Approximation (COBYLA) algorithm and the Nelder-Mead algorithm.

The COBYLA algorithm is a method for solving limited optimisation problems without using derivatives, based on Powell's approach of approximating the problem with linear functions [27]. The Nelder-Mead algorithm was chosen for this 6 operations on a simplex, iteratively adjusting its shape to minimize the objective function [28].

D. Heuristic Methods

1 The Cyclic Coordinate Descent

The CCD algorithm was chosen as the greedy search method. It iterates towards an approximate solution, by taking advantage of the geometric configuration of the robot. Unlike other algorithms that try to optimize all the joint angles at once, it tries to optimize each joint angle one at a time before moving on to the next joint.

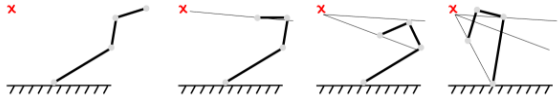

**Figure 5**. Illustration of the CCD method.

Each joint is optimized one at a time till a threshold criterion is met.

Due to the cyclic nature of CCD, the convergence graph is very jittery and almost unusable. This noise was smoothened by clipping the peaks every time a new minimum value was set during the cyclic phases.

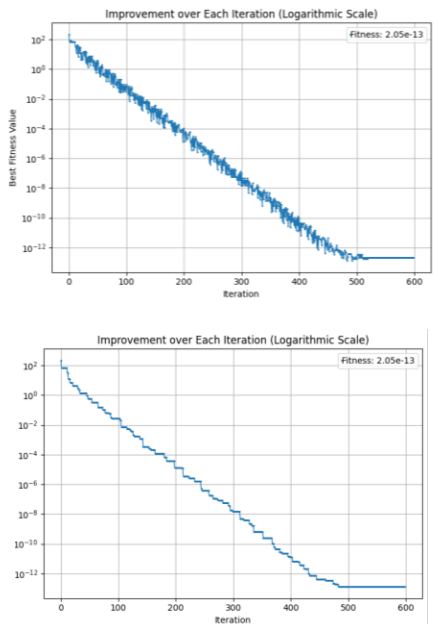

**Figure 6**. CCD before clipping and after clipping.

### 2. Simulated Annealing (SA)

Simulated Annealing is inspired by the annealing process in metallurgy. For the inverse kinematics solution, a random set of joint angles is initially generated and passed through a high amount of perturbance T. This generates a number of potential candidates, L. Each candidate is compared to the previous one based on its fitness and a selection is made using a probability that is proportional to the current "temperature" and the fitness difference. The iteration is then repeated with smaller values of T until a tolerance or maximum iteration count is met.

---

**Algorithm 1**. Pseudocode for SA

1: fitness: TargetDistance td()
2: **Input**: Target (x,y,z)
3: **Output**: JointAngles ($q_{1-7}$)
4: **procedure** SA
5:   **Input**: Tmax, Tmin, cooling_rate cr
6:   **Output**: Best $q_{1-7}$
7:   **Initialize**: rand($q_{1-7}$), Enow = fitness($q_{1-7}$)
8:   **while** Tnow > Tmin do
9:     **for** q in $q_{1-7}$ **do**
10:      Select a new angle qnew = qn + $\Delta$q
11:      Enew = fitness(qnew)
12:      $\Delta$E = Enew - Enow
13:      **if** $\Delta$E < 0 or rand() < $e^{(\Delta E/Tnow)}$ **then**
14:        Accept qnew
15:        Update Enow = Enew
16:      **end if**
17:    **end for**
18:    Decrease temperature: Tnow = Tnow *cr
19:  **end while**
20:  **Return** Best $q_{1-7}$
21: **end procedure**

### E. Evolutionary Algorithms

#### 1. Genetic Algorithm (GA)

GA is inspired by the principles of natural selection and genetics. It mimics the process of evolution to iteratively search for optimal or near optimal solutions. For finding the inverse kinematics, each individual represents a set of joint angles (7 genes) for the robot arm. GA comprises of evaluation, selection, crossover, mutation and repopulation.

---

**Algorithm 2**. Pseudocode for GA

1: **fitness function**: TargetDistance td()
2: **Input**: Target (x,y,z)
3: **Output**: JointAngles (q1,q2,q3,q4,q5,q6,q7)
4: **procedure** GA
5:   **Input**: Pop, rand($q_{1-7}$), CrossPr(cp) , MutPr(mp)
6:   **Output**: Best $q_{1-7}$ in all generations
7:   **Initialize**: $Pop_{size}$*RandAngles
8:   **for** each $q_{1-7}$ **do**
9:     Compute td()
10: **end for**
11: **repeat**
12:   Select parents $p_1$, $p_2$ based on td()
13:   **for** all new children do
14:     crossover $p_1$, $p_2$
15:     mutate each new joint angle with Pr(mp)
16:   **end for**
17:   Check td() of new $q_{1-7}$
18:   Replace worst performing old with best new
19: **until** error<threshold or max_iterations
20: **end procedure**

## 2. Differential Evolution (DE)

Differential Evolution (DE) is a specific subset of genetic algorithms, focusing on real-valued vector optimization problems like Inverse Kinematics. Here, the genotype represents a set of real-valued vectors. During the mutation and crossover operations, DE utilizes the differences between two or more vectors in the population. This process involves creating a new vector by adding a random proportion of the difference to one of the existing vectors, along with a small amount of random noise. By iteratively applying these operations across the population, DE aims to evolve towards optimal or near-optimal solutions within the search space.

## F. Swarm Intelligence

### 1. Particle Swarm Optimization (PSO)

PSO is inspired by the social behaviour of bird flocks or fish schools. The PSO algorithm initializes a population of potential solutions (particles), each representing a set of joint angles. These particles move through the solution space, adjusting their positions based on their own best-known solution and the swarm's best-known solution.

| **Algorithm 3**. Pseudocode for PSO |
| --- |
| 1: fitness: TargetDistance td() |
| 2: **Input**: Target (x,y,z) |
| 3: **Output**: JointAngles ($q_{1-7}$) |
| 4: **procedure** PSO |
| 5:   **Input**: num_particles, max_iter, c1, c2, w |
| 6:   **Output**: Best $q_{1-7}$ |
| 7:   **Initialize**: particles with random q1-7 and velocities |
| 8:   **Initialize**: pbest for each particle |
| 9:   **Initialize**: gbest among all particles |
| 10:   **for** iter = 1 to max_iter **do** |
| 11:     **for** each particle in particles **do** |
| 12:       Update v = w*v + c1*rand()*(pbest - q) + c2*rand()*(gbest - q) |
| 13:       Update position: q = q + v |
| 14:       Evaluate fitness: td(q) |
| 15:       **if** td(q) < td(pbest) **then** |
| 16:         Update pbest: pbest = q, td(pbest) = td(q) |
| 17:       **end if** |
| 18:       **if** fitness(q) < fitness(gbest) **then** |
| 19:         Update gbest: gbest = q, td(gbest) = td(q) |
| 20:       **end if** |
| 21:     **end for** |
| 22:   **end for** |
| 23:   **return** gbest |
| 24: **end procedure** |

The position and velocity equations of a particle i in the d-dimensional search space at time t are given by:

$$x_i(t+1) = x_i(t) + v_i(t+1) \quad (10)$$

$$v_i(t+1) = w \cdot v_i(t) + c_1 \cdot r_1 \cdot (pbest_i - x_i(t)) + c_2 \cdot r_2 \cdot (gbest - x_i(t)) \quad (11)$$

Where:

- $x_i(t)$ is the current position of the particle.
- $v_i(t+1)$ is the velocity of the particle at time t+1.
- w is the inertia weight
- $c_1$ and $c_2$ are the acceleration coefficients.
- $r_1$ and $r_2$ are random values between 0 and 1
- $pbest_i$ is the personal best of particle i.
- $gbest$ is the global best of all the particles.

### 2. Quantum Particle Swarm Optimization

Quantum Particle Swarm Optimization (QPSO) incorporates concepts from quantum computing to classical PSO. QPSO aims to enhance the exploration and exploitation capabilities of the PSO algorithm by introducing quantum behaviours into the particle movement and update rules.

### 3. Artificial Fish Swarm Algorithm (AFSA)

AFSA is inspired by the nature of fish swarms. An initial population of fish is first initialized with random positions representing potential joint configurations. Fish with better fitness values attract others towards them, mimicking the concept of prey fish being attracted to areas with abundant food. It is characterized by 3 behaviours: preying, swarming, and following where each Artificial Fish is attracted to the next using these behaviours.

## E. Machine Learning Models

The application of machine learning to solve inverse kinematics, particularly in the context of higher degrees of freedom, is an intriguing proposition. The approach is especially valuable in cases where the constraints are complex, and the fitness function is difficult to determine.

The models employed in this study were trained using a dataset consisting of 2,097,157 rows. Each row is an array of 10 items which are the joint angles and their corresponding positions.

$$(\theta_1, \theta_2, \theta_3, \theta_4, \theta_6, \theta_7, x, y, z)$$

The dataset was generated in an iterative way by looping through joint angles in a sequential manner to avoid conflicts. A small degree of random noise between ±0.1 rad was added to each angle so the dataset was not perfectly smooth.

The dataset was divided into a 25% testing set and the remaining data was used for training.

After training, 10,000 random samples were generated, and the average fitness of the dataset was recorded.

When the model predicts negative joint angles and the actual joint angles are also negative, the model's predictions might not be improving upon a naive baseline (like predicting the mean). Since normalization was challenging, an absolute value of $r^2$ was recorded to account for angle wrapping.

### 1. Regression

Regression techniques like linear regression, polynomial regression and neural networks are used to create hyperplanes that represent the relationship between inputs and outputs. For a simple linear regression with one input variable (x), the predicted output (y) given a slope (m) and an intercept (b) is given by:

$$y = mx + b \qquad (12)$$

For a 7 DOF robot arm, a multiple linear regression is more applicable.

$$y = b_0 + b_1\theta_1 + b_2\theta_2 + b_3\theta_3 + b_4\theta_4 \qquad (13)$$
$$+ b_5\theta_5 + b_6\theta_6$$
$$+ b_7\theta_7$$

- Where $b_0$ is the y-intercept.
- $\theta_n$ is the joint angle at position n.
- $b_n$ are the coefficients for each position n.

For a polynomial regression, the general equation for a polynomial of degree n is:

$$y = b_0 + b_1\theta + b_2\theta^2 + \cdots + b_n\theta^n \qquad (14)$$

Both linear and polynomial regression models were created. A polynomial degree of 8 was used after several tests. The polynomial degree for other values is recorded in the Appendix 1.

### 2. Decision Trees

Decision trees are adept at capturing relationships between non-linear equations like these since they partition the input space into regions, where each region corresponds to a specific output value. They mimic human decision-making by creating a tree-like model of decisions based on features of the dataset.

The Decision Tree partitions the joint space into regions, with each leaf node representing a set of joint angles.

The final depth of the decision tree was 84 layers.

### 3. Feed Forward Neural Network

The feed forward network used is made of 13 layers. The input layer consists of 3 inputs corresponding to the x,y and z target positions. The output layer consists of 7 outputs corresponding to the joint angle positions. The training was run for 100 epochs.

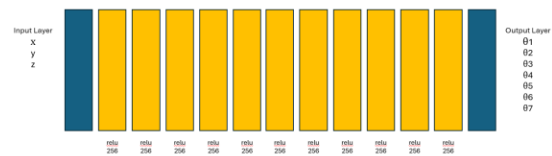

**Figure 7**. NN Architecture

### F. Novel Solution (DTNR)

A novel hybrid method is proposed which combines the Decision Tree + Newton Raphson method (DTNR). It combines the speed of the Decision Tree with the precision and accuracy of the Newton-Raphson method. It does it by using the output of the decision tree as the initial guess for the Newton method. The numerical stage only computes for the first 3 joints.

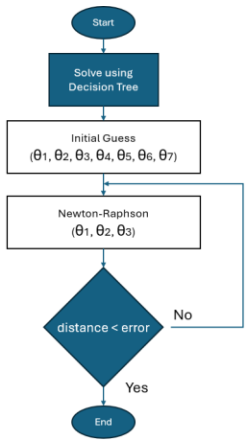

Figure 8. Flowchart for proposed DTNR method

## III. Results

A. Hyperparameter Tuning

Unless stated otherwise, time curves are shown in green and fitness curves are shown in blue.

1. Genetic Algorithm

The parameters that need to be optimized for the GA are the population size and the mutation rate. The time taken and best fitness value were plotted against increasing number of population sizes and mutation probability.

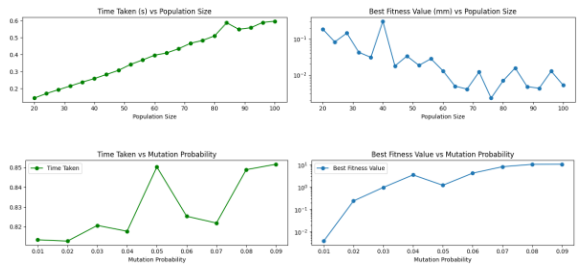

Figure 9. GA Tuning Graphs

From the results, a population size of 30, with a mutation probability of 1% was chosen as the optimal parameters for accuracy and speed.

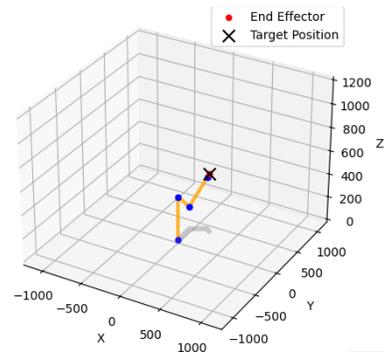

Figure 10. GA 3D plot with tuned parameters.

2. Differential Evolution

For DE, the minimum time taken was plotted against the average 2 of the best 5 attempts for each fitness, successive population size steps. Despite this, the fitness trend was still chaotic, and this shows that an increase in population size does not have a significant correlative effect with error minimization. A population size of 20 was chosen with a mutation probability of 0.1%.

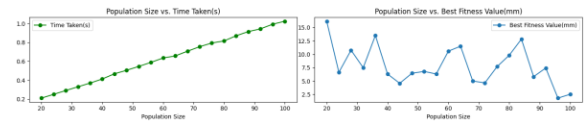

Figure 10. DE tuning graphs.

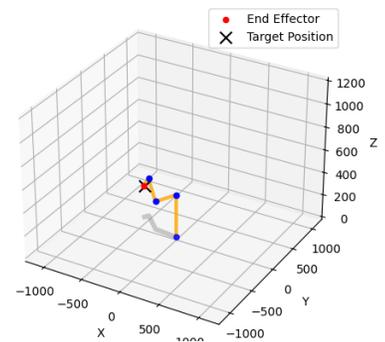

Figure 11. DE 3D plot with tuned parameters

3. Simulated Annealing

The parameters that need to be optimized for the Simulated Annealing function are the maximum temperature ($T\_max$), minimum temperature ($T\_min$) and the maximum number of attempts to find a better solution in the neighbourhood (MSC). Each node was run 5 times and the best fitness and average of the best 2 times were plotted.

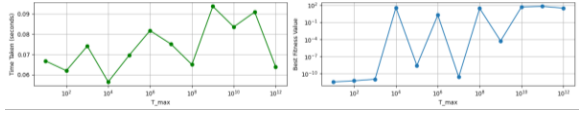

**Figure 11**. SA Tmax vs time and fitness

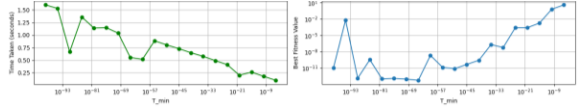

**Figure 12**. SA Tmin vs time and fitness

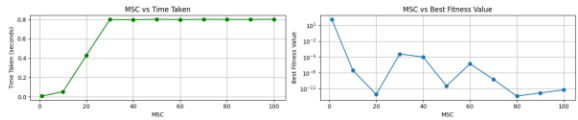

**Figure 13**. MSC vs time and fitness

**Table 2**. SA parameters

| T_max | **100** |
|---|---|
| T_min | 1e-50 |
| Max Stay Counter | 20 |

4. Particle Swarm Optimization

For the PSO, an increase in population size comes with an increase in the time take, error rate and a general reduction in iteration count. A population size of 20 was chosen.

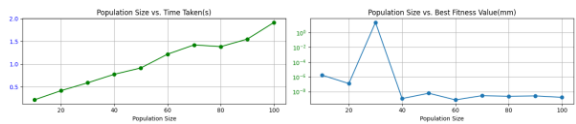

**Figure 15**. PSO Parameter graph

5. Artificial Fish Swarm Algorithm (AFSA)

The parameters that need to be optimized are the population size, exploration probability (q), Maximum Attempt Size (MAS), and visual range (delta)

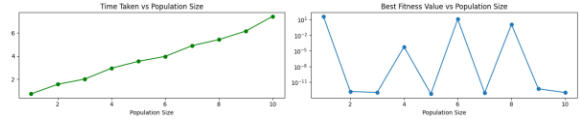

**Figure 16**. AFSA Time Taken and Fitness vs Population.

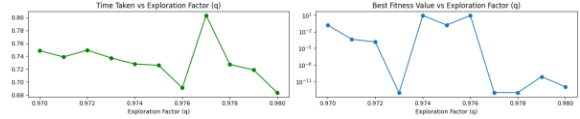

**Figure 17**. AFSA Time and Fitness vs Exploration Factor (q)

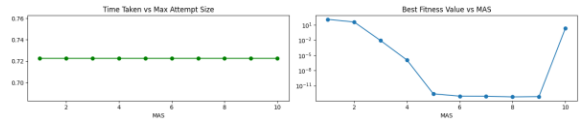

**Figure 18**. AFSA Time and fitness vs MAS

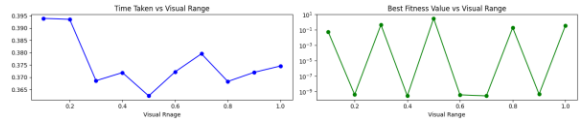

**Figure 19**. AFSA Time and fitness vs Visual Range

**Table 3**. AFSA parameters

| Population Size | **1** |
|---|---|
| q | 0.971 |
| MAS | 4 |
| Visual Range | 0.6 |

B. Experimental Results

1. Computational Specifications

All tuning, training, and testing operations were performed on a single DELL 15 PC with AMD Ryzen 5 processor. The code was written in python and the TensorFlow library was used for the machine learning operations.

2. Machine Learning Results

The machine learning methods yielded unsatisfactory results.

**Table 4.** ML Results from 100,000 randomly generated samples

| Model | $r^2$ on validation | MSE | Average Fitness (mm) |
|---|---|---|---|
| Linear Regression | 0.00752 | 3.2210 | 782 |
| Polynomial Regression (8 degrees) | -0.03298 | 3.3524 | 719 |
| Decision Tree | -0.76185 | 5.7142 | 24 |
| Feed Forward NN | 0.09713 | 2.9307 | 665 |

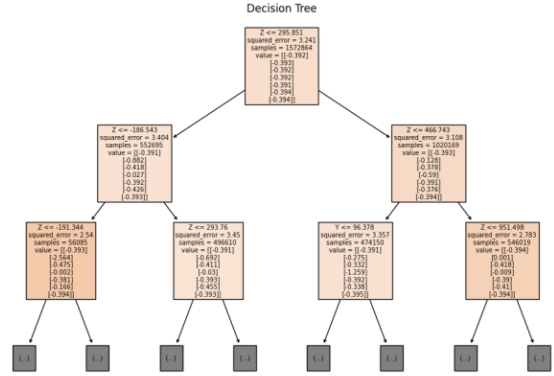

**Figure 20.** Decision Tree Diagram of the first 3 layers of the model

The Decision tree resulted in $R^2$ of -0.76185 which suggests overfitting, as decision trees are prone to such problems. Despite this, 10,000 new random positions were generated, and the decision tree model performed the best out of all the machine learning methods that were tested. It is noted that this might be attributed to angle wrapping. The final depth of the decision tree was 74 layers.

3. Overall Performance Results

A set of 100 random target positions were generated and consistently applied for all algorithms. A fitness score of < 1mm was used as the threshold which would be used to determine the success rate (SR). The weighted average was used as the overall best fitness in the successful cases. The data was then used to plot the average iteration graphs with the Standard Deviation (SD).

**Table 5.** Algorithm Performance.

| Algorithm | Iteration Count | Best Fitness | Worst Fitness | Best Time (s) | Worst Time (s) | Average Fitness (mm) | Average Fitness Weighted | SD | Average Time (s) | Success Rate |
|---|---|---|---|---|---|---|---|---|---|---|
| **DTNR** | 15 | 1.00e-15 | 1051 | 0.0010 | 0.0137 | 7.18e-14 | 5.31e-15 | 232.3 | 0.0075 | 84 |
| **NR** | 30 | 1.00e-15 | 424 | 0.0099 | 0.0595 | 1.08e-12 | 1.71e-14 | 42.20 | 0.0279 | 99 |
| **NM** | 799 | 1.00e-15 | 84.5 | 0.1172 | 0.3080 | 0.01426 | 4.54e-15 | 9.454 | 0.1648 | 94 |
| **SA** | 321 | 1.00e-15 | 36.8 | 0.1397 | 2.9963 | 0.01207 | 5.95e-14 | 5.440 | 2.2358 | 92 |
| **PSO** | 20 | 1.00e-15 | 71.9 | 1.1676 | 2.5034 | 0.00041 | 1.57e-15 | 10.96 | 1.5648 | 94 |
| **QPSO** | 302 | 1.00e-15 | 13.9 | 0.8130 | 1.1414 | 0.02350 | 8.95e-15 | 2.084 | 0.8720 | 84 |
| **CCD** | 300 | 4.11e-8 | 213.2 | 0.0462 | 199.9 | 0.03000 | 2.06e-7 | 49.95 | 5.4192 | 41 |
| **AFSA** | 41 | 5.73e-14 | 212.8 | 0.7266 | 1.010 | 0.06696 | 4.43e-13 | 41.32 | 0.8155 | 47 |
| **GA** | 100 | 0.00011 | 0.778 | 21.185 | 53.17 | 0.01764 | 0.00079 | 0.1058 | 22.751 | 100 |
| **DE** | 100 | 0.00395 | 1.132 | 21.521 | 28.85 | 0.14773 | 0.05484 | 0.1489 | 22.565 | 99 |

## DTNR

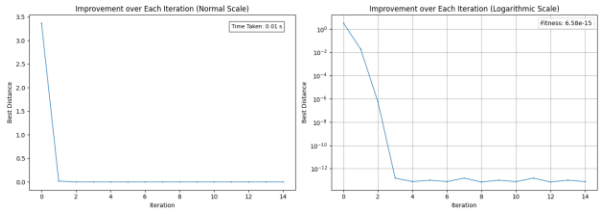

## DE

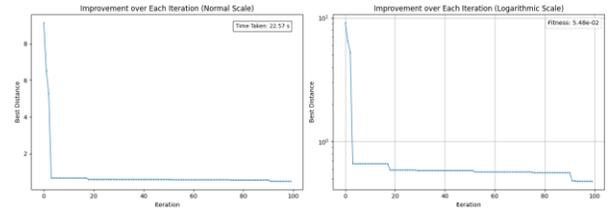

## Newton-Raphson

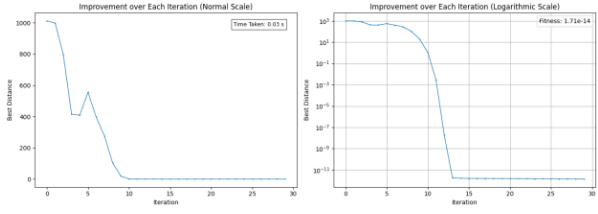

## SA

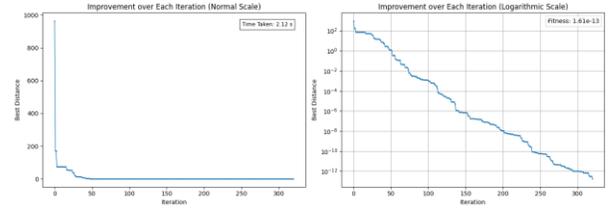

## Nelder Mead

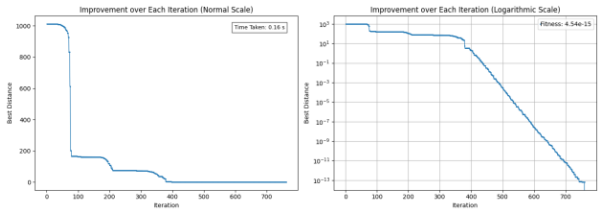

## PSO

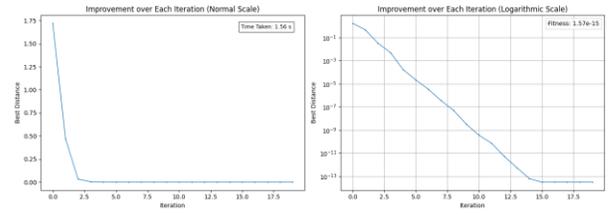

## CCD

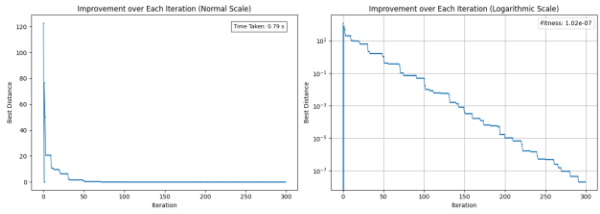

## QPSO

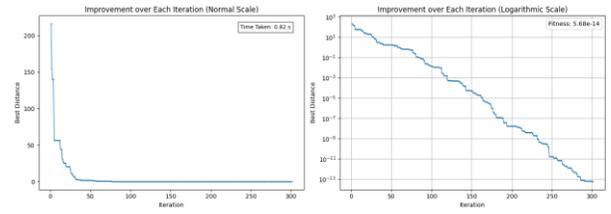

## GA

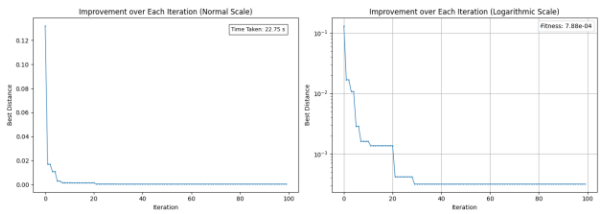

## AFSA

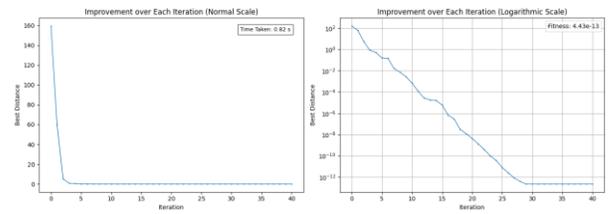

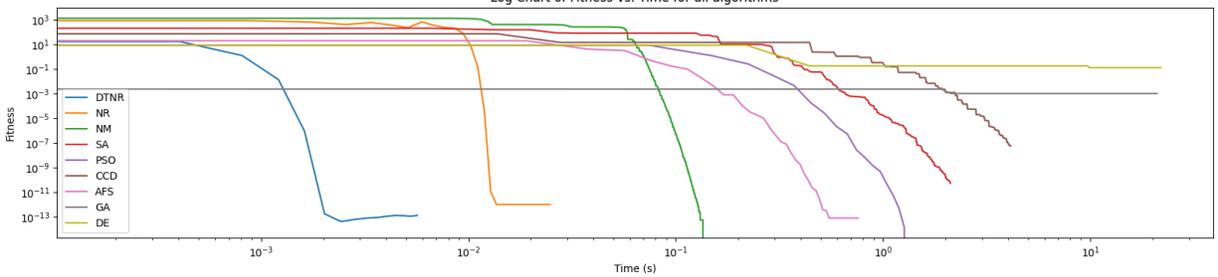

4. Results from the table

- The DTNR method is the fastest algorithm, being 3.7 times faster than its closest competitor which is the NR method, and 208 times faster than PSO. It also shows the worst fitness when wrong.
- The NR shows very high stability with a SR of 99% and speed and is the most preferred choice. It is closely followed by the SA due to the SA's greater solution diversity.

5. Average Convergence Graphs

The iteration graphs are not from a single test but from the average of the successful runs. Since the time remained within a consistent range, a simple average was used for the overall time. However, the final fitness score was calculated from the weighted averages from the successful iterations.

The final iteration graph gives us a comparative understanding of the performance. Trends towards the top right perform worse than those towards the bottom left.

### IV. Conclusions

The complexities of finding the IK solution for redundant serial manipulators was explored. A 7-DOF robot arm was analysed. 14 optimization techniques were tested including a novel DTNR method. The following was concluded.

- The DTNR method emerges as the top performer in terms of both best and worst time efficiency among the algorithms tested. It is 4 times faster than the fastest algorithm tested and 244 times faster than PSO. Unfortunately, it has a poor reliability of 84%, only surpassing that of the CCD and AFSA.
- The evolutionary algorithms, albeit slower, demonstrate a consistent ability to converge to a solution. Furthermore, they offer a broader range of solutions once convergence is achieved.
- AFSA and CCD stand out as the least effective performers, primarily due to their low success rates. Notably, CCD ranks lowest overall, mainly attributed to its propensity for entrapment within iterative loops, as evidenced by its worst time metrics.
- The machine learning methods performed poorly in general.

Recommendations

- The standard deviation of the DTNR could be improved, making the model significantly more reliable.

# Appendix

Appendix 1. Polynomial Values at different degrees

| Degree | $r^2$ | MSE | Average Fitness |
|---|---|---|---|
| 4 | 0.04502 | 3.0975 | 683 |
| 5 | 0.04531 | 3.0965 | 684 |
| 6 | 0.05501 | 3.0651 | 667 |
| 7 | 0.05265 | 3.0727 | 652 |
| **8** | **0.03326** | **3.1356** | **650** |
| 9 | 0.04330 | 3.1031 | 690 |
| 10 | -0.5960 | 5.1749 | 698 |

Appendix 2. Some randomly generated target positions from the experiments

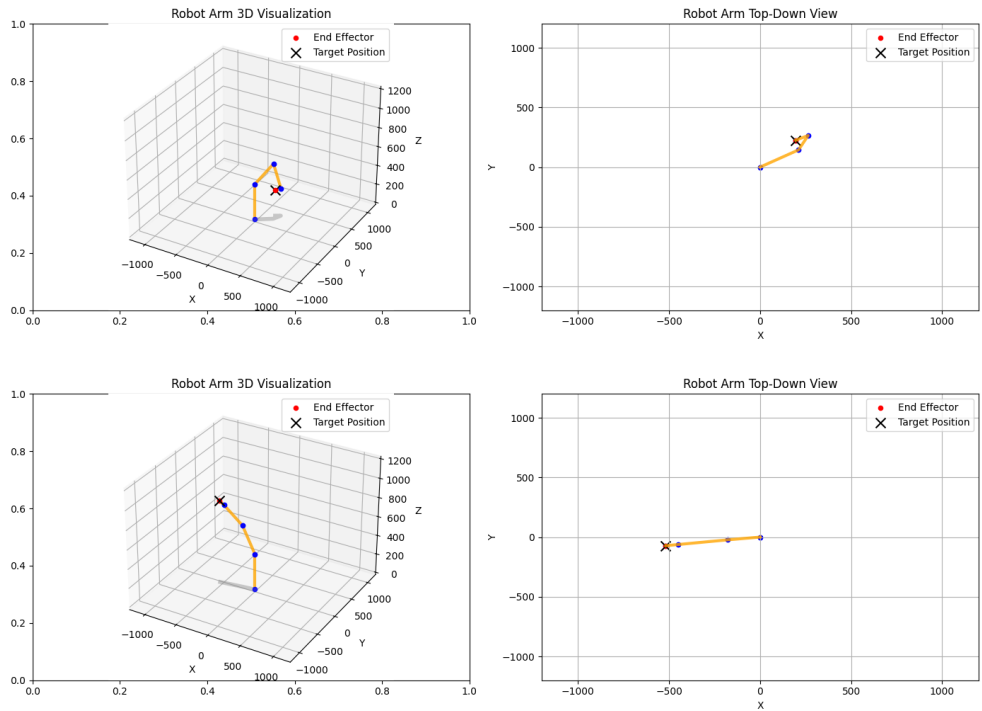